\documentclass[runningheads]{llncs}
\usepackage[T1]{fontenc}
\usepackage{graphicx,verbatim}
\usepackage[table]{xcolor}
\definecolor{mygray}{gray}{.92}
\usepackage{algorithmic}
\usepackage{textcomp}
\usepackage{marvosym}
\usepackage{graphicx}
\usepackage{subfigure}
\usepackage{multirow}
\usepackage{bm}
\usepackage{mathrsfs}
\usepackage{amsmath}
\usepackage{amsfonts,amssymb} 
\usepackage{graphicx,epstopdf,subfigure}
\usepackage{makecell}
\usepackage{pifont}
\usepackage[colorlinks=true, linkcolor=blue, citecolor=green]{hyperref}
\usepackage{makecell}
\usepackage{booktabs}
\usepackage{pifont}
\usepackage{float}
\usepackage{lineno}
\usepackage{tikz}
\usetikzlibrary{positioning} 
\usepackage{lmodern}    
\usepackage{mweights}   

\usepackage{makecell} 
\usepackage{array} 

\begin{document}
%
\title{Medical Large Vision Language Models with Multi-Image Visual Ability}
%

\author{Xikai Yang\inst{1} \and Juzheng Miao\inst{1} \and Yuchen Yuan\inst{1} \and Jiaze Wang\inst{1} \and Qi Dou\inst{1,2} \and Jinpeng Li\inst{1}\textsuperscript{*} \and Pheng-Ann Heng\inst{1,2}}  
\authorrunning{X. Yang et al.}
\institute{Dept. of Computer Science and Engineering, The Chinese University of Hong
Kong, Hong Kong, China \and
Institute of Medical Intelligence and XR, The Chinese University of Hong Kong,
Hong Kong, China \\
\email{jpli21@cse.cuhk.edu.hk}}

\maketitle              
\begin{abstract}
Medical large vision-language models (LVLMs) have demonstrated promising performance across various single-image question answering (QA) benchmarks, yet their capability in processing multi-image clinical scenarios remains underexplored.
Unlike single image based tasks, medical tasks involving multiple images often demand sophisticated visual understanding capabilities, such as temporal reasoning and cross-modal analysis, which are poorly supported by current medical LVLMs.
To bridge this critical gap, we present the Med-MIM instruction dataset, comprising 83.2K medical multi-image QA pairs that span four types of multi-image visual abilities (temporal understanding, reasoning, comparison, co-reference).
Using this dataset, we fine-tune Mantis and LLaVA-Med, resulting in two specialized medical VLMs: MIM-LLaVA-Med and Med-Mantis, both optimized for multi-image analysis. Additionally, we develop the Med-MIM benchmark to comprehensively evaluate the medical multi-image understanding capabilities of LVLMs. We assess eight popular LVLMs, including our two models, on the Med-MIM benchmark.
Experimental results show that both Med-Mantis and MIM-LLaVA-Med achieve superior performance on the held-in and held-out subsets of the Med-MIM benchmark, demonstrating that the Med-MIM instruction dataset effectively enhances LVLMs' multi-image understanding capabilities in the medical domain. 
The Med-MIM instruction dataset, benchmark, and fine-tuned models will be available at \href{https://github.com/Xikai97/Med-MIM}{Med-MIM}.


\end{abstract}

\section{Introduction} \label{introduction}

Large Vision-Language Models (LVLMs), with proper visual instruction tuning, such as LLaVA~\cite{liu2024visual}, have demonstrated remarkable potential in various image perception and QA tasks. To further adapt LVLMs to the medical domain, recent developments have introduced medical-specific LVLMs~\cite{wu2023towards,zhang2023pmc,li2024llava,bai2024m3d}, such as LLaVA-Med~\cite{li2024llava} with impressive diagnostic capabilities.
However, these models are typically designed for single-image based tasks, lacking instruction tuning tailored for multi-image question answering. As a result, their performance is often limited when addressing medical multi-image tasks, which are more complex yet crucial in the field of medical image analysis.
Unlike natural images, medical multi-image is often closely tied to clinical needs. During the diagnostic process, doctors typically analyze multi-visit, multi-view, and multi-modality data in a systematic manner to produce a comprehensive and accurate assessment~\cite{korot2021code,poulakis2022multi}. 
For instance, given screening data from two consecutive visits of a patient, the goal might be to determine whether the patient’s condition is worsening~\cite{rousan2020chest}. In retrospective studies, such as monitoring Alzheimer’s disease, the question becomes whether we can predict a patient’s future cognitive status using his longitudinal neuroimaging data~\cite{johnson2009longitudinal}. More commonly, when working with multiple imaging modalities, such as MRI from various sequences, the challenge is to integrate information from diverse sources to provide a reliable diagnosis~\cite{alleman2023multimodal,calhoun2016multimodal}.

While some pioneering works have started to explore the performance of LVLMs in natural multi-image settings~\cite{meng2024mmiu,Jiang2024MANTISIM,liu2406mmdu}, current medical LVLMs still overlook the development of multi-image visual capabilities.
Due to the challenges associated with collecting and processing medical multi-image data, only a limited number of works attempted to develop multimodal large models trained on such data.
Yang \textit{et al.}~\cite{jinxia2024medST} pretrained a medical vision-language framework to leverage spatial and temporal relationships across multiple X-ray views and historical records. 
However, their work primarily focused on multimodal alignment and pretraining, without providing the capability for open-ended visual question answering.
Additionally, their dataset was restricted to chest X-ray images, limiting its applicability to other medical imaging scenarios.
Other studies~\cite{srivastavprocedure,seyfioglu2024quilt} have explored video-language models, achieving strong performance in analyzing surgical scenes and histopathology videos. Nevertheless, these methods are tailored to specific video domains and fail to address broader multi-image tasks, such as longitudinal temporal forecasting and multi-view understanding, which remain unexplored.
Furthermore, existing medical VQA benchmarks, like VQA-RAD~\cite{lau2018dataset}, do not adequately assess multi-image comprehension, leaving a critical gap in evaluating models' clinical multi-image visual understanding capabilities.

To enhance multi-image visual abilities in medical LVLMs, we present the \textbf{Med}ical \textbf{M}ulti-\textbf{IM}age (Med-MIM) instruction dataset, along with the Med-MIM benchmark, designed for comprehensively evaluating LVLMs across various types of multi-image visual tasks. To the best of our knowledge, both the Med-MIM instruction dataset and Med-MIM benchmark represent the first focused effort on medical multi-image analysis.
Our key contributions are summarized as follows:
(1) We construct the Med-MIM instruction dataset, which contains 83.2K medical multi-image instruction samples. The constructed dataset is organized into four subsets designed to enhance VLMs' co-reference, comparison, reasoning, and temporal understanding abilities. 
(2) We establish the Med-MIM benchmark, which includes a held-in evaluation subset derived from the Med-MIM instruction dataset. Additionally, we construct two held-out multi-image datasets to evaluate the models' generalized zero-shot multi-image visual capabilities.
(3) Using the Med-MIM instruction dataset, we develop two specialized medical VLMs: MIM-LLaVA-Med and Med-Mantis, tailored for multi-image analysis. We compare these models against several state-of-the-art VLMs, and experimental results demonstrate that our models exhibit superior multi-image visual capabilities on both the held-in and held-out Med-MIM benchmarks.

\section{Method} \label{method}
\begin{figure}[t]
    \centering
	\includegraphics[width=0.90\textwidth]{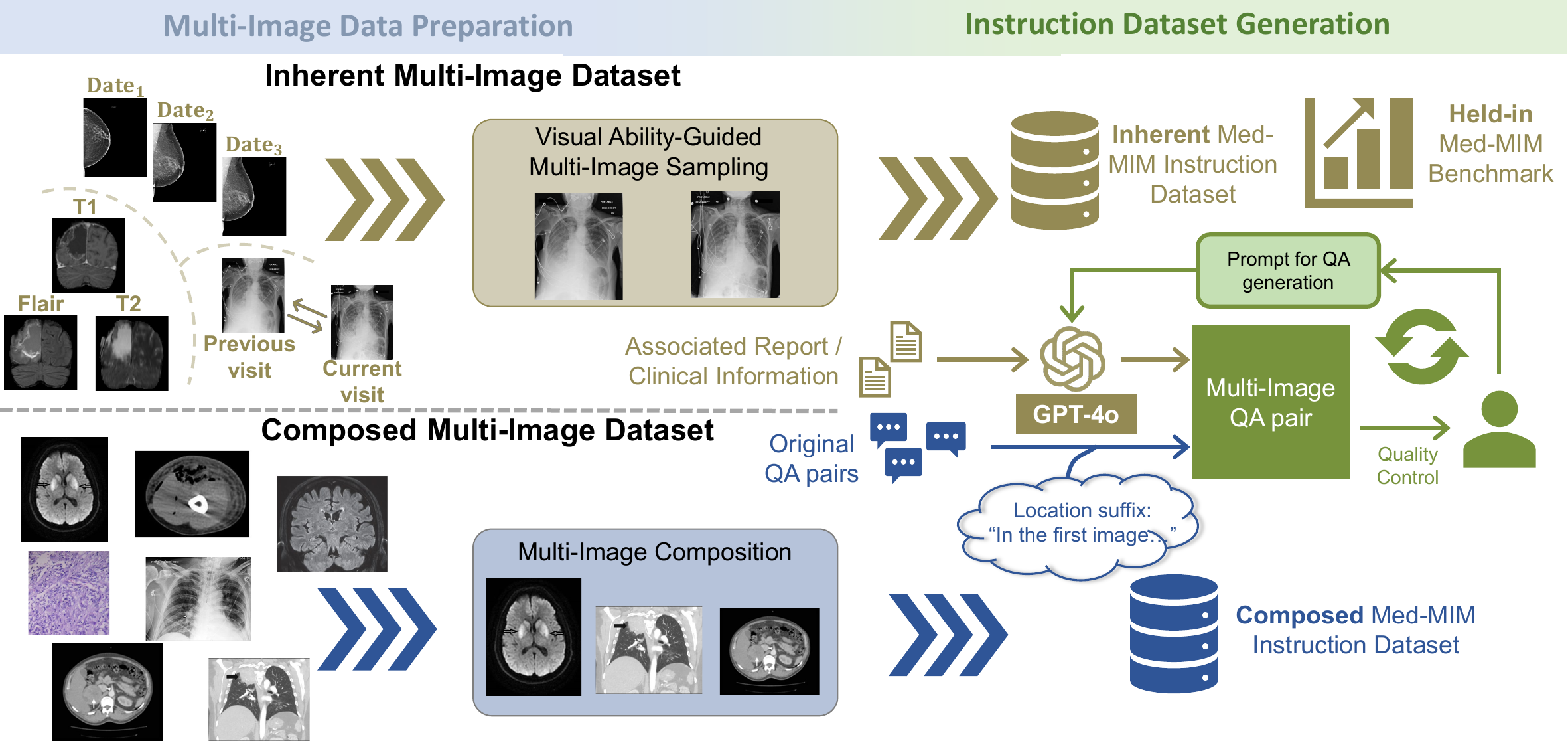}
	\caption{
    Illustration of the pipeline to generate Med-MIM dataset and benchmark. 
    } 
    \label{fig:data_generate}
\end{figure}

\begin{figure}[t]
    \centering
    \includegraphics[width=0.95\textwidth]{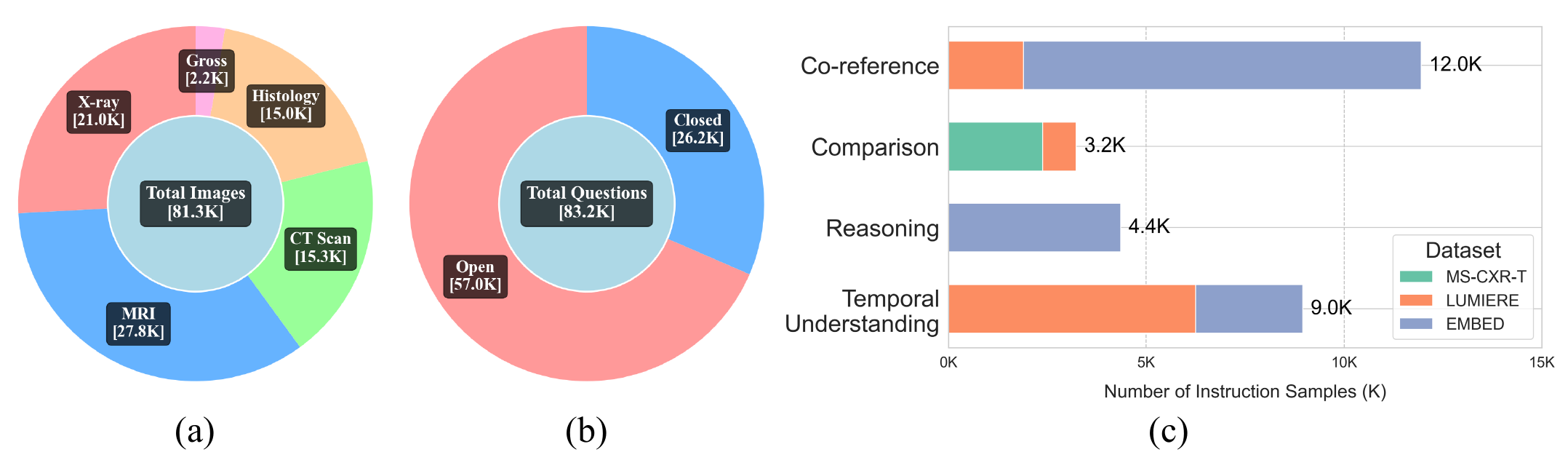}
    \caption{Comprehensive analysis of the Med-MIM dataset: (a) Image modality composition; (b) Distribution of the question types; (c) \textbf{Inherent} multi-image dataset distribution categorized by four core visual abilities.} 
    \label{fig:instruct_statistic}
\end{figure}

\begin{figure}[t]
    \centering
	\includegraphics[width=0.95\textwidth]{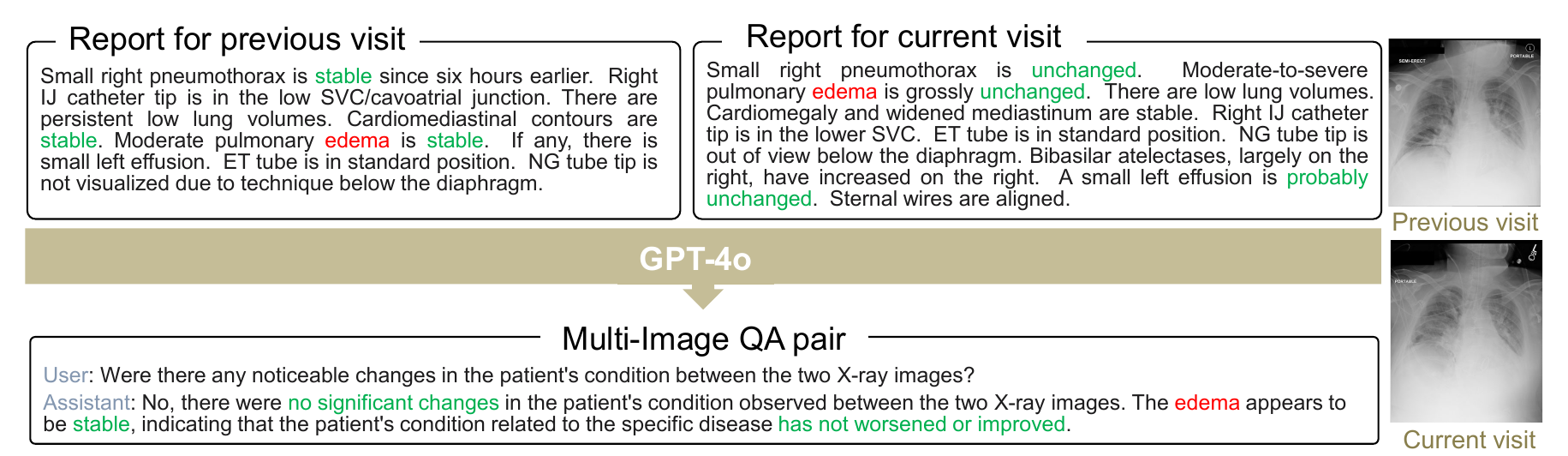}
	\caption{Example of generated multi-image QA pair in the MS-CXR-T dataset.} 
    \label{fig:qa_example}
\end{figure}

We create the Med-MIM Instruction dataset and benchmark following two steps: multi-image data preparation and instruction dataset generation (as illustrated in Fig.\ref{fig:data_generate}). During the data preparation phase, two types of multi-image samples (inherent and composed) are collected. Subsequently, GPT-4o~\cite{hurst2024gpt}, combined with iterative refinement, is employed to generate the inherent Med-MIM instruction dataset and the held-in benchmark. Meanwhile, the composed Med-MIM instruction dataset is created by appending location-specific suffixes to the original QA pairs, ensuring alignment with multi-image visual understanding.
\\
\textbf{Dataset Collection.}
Fig.\ref{fig:instruct_statistic} presents an overview of our constructed Med-MIM instruction dataset composed of 83.2K medical multi-image QA instruction samples from five main domains.
Unlike previous medical instruction dataset like PMC-VQA~\cite{zhang2023pmc} or LLaVA-Med VQA~\cite{li2024llava} that primarily based on published literature, we construct the Med-MIM instruction dataset from two sources: inherent multi-image datasets and composed multi-image dataset, where inherent multi-image datasets contain naturally occurring multi-image scenes such as longitudinal inspections, multi-modality or multi-view grounded diagnoses, etc. 
To curate high-quality multi-image dataset, we group multiple images based on tasks related to various visual abilities, utilizing three large-scale inherent multi-image datasets: MS-CXR-T~\cite{bannur2023ms}, EMBED~\cite{jeong2023emory}, and LUMIERE~\cite{suter2022lumiere}.
All three datasets include multi-visit images arranged in chronological order, while EMBED and LUMIERE further incorporate multi-view and multi-modality data.
Additionally, to further enhance the richness of the dataset and include a broader variety of medical images, we construct a composed multi-image dataset by manually grouping multiple images from the LLaVA-Med VQA~\cite{li2024llava} dataset. 
We cap the length of multi-image sequences at a maximum of three, allowing for grouping more multi-image QA pairs while ensuring feasibility with VLM fine-tuning under limited computational resources.
\\
\textbf{Four Multi-image Visual Abilities.} 
Following the MANTIS~\cite{Jiang2024MANTISIM}, we can divide all collected multi-image samples into four categories corresponding to four different visual abilities (Fig.\ref{fig:instruct_statistic} (c)). However, for medical multi-images, more specific tasks can be linked to various abilities as follows. 
\\
\textbf{\textit{ (1) Temporal Understanding}}: We focus on multi-visit data and construct QA pairs following the temporal forecasting task. For example, ``Given the first two consecutive visits collected at \{date1\} and  \{date2\}, will the subject become diseased in the next visit?'' We filter 9K temporal understanding related samples from the LUMIERE and EMBED datasets.
\\
\textbf{\textit{(2) Reasoning}}: We restrict reasoning-type multi-image tasks to multi-view diagnosis in the EMBED dataset. A total of 4.4K samples are collected. The LVLM agent will be asked to diagnose patient from both two views (Cranio-Caudal \& Medio-Lateral Oblique), such as ``Considering the features observed in the two provided mammographic views, what is the BIRADS category for the findings?''
\\
\textbf{\textit{(3) Comparison}}: By comparing multiple images, we aim to find the similarities or differences. In our instruction dataset, we obtain comparison-type multi-image samples from both the MS-CXR-T and LUMIERE datasets. Note that although MS-CXR-T contains temporal relations, the actual task is to determine progression status based on previous and current visits, which requires the model to find differences between two visits. For instance, ``Could you explain the differences between the X-ray images from the first visit and the second visit?''
\\
\textbf{\textit{(4) Co-reference}}: We construct a co-reference-related subset from both the LUMIERE and EMBED datasets by asking the model questions such as, ``Which image, the first or the second, represents the view that captures the breast from above?'' By incorporating location into the question, we aim to enable the VLM to understand both image location and content simultaneously. Additionally, for the composed multi-image dataset, we categorize it into the co-reference subset as well. To encode proper location information, the original single-image QA pairs are updated by adding prefixes such as ``In the first image'' or ``In the second image.'' (as illustrated in Fig.\ref{fig:data_generate}).
\\ 
\textbf{Instruction Dataset Generation.}
After collecting the multi-image sets, we integrate the textual data for each image and generate the corresponding question-answer pairs.
Unlike the composed Med-MIM instruction dataset, which can be created by simply appending location-specific suffixes (e.g., ``in the first image'') to the original QA pairs, the inherent Med-MIM dataset requires a more sophisticated approach. For this, we prompt language-only GPT-4o to generate multi-image QA pairs by leveraging the medical reports or clinical information associated with each image. We also incorporate human reviewers as a quality control measure to iteratively refine the generated results, ensuring both accuracy and clarity.
Fig.\ref{fig:qa_example} shows one multi-image QA example generated by GPT-4o based on the medical reports from two visits in the MS-CXR-T dataset.
\\
\textbf{Multi-image Visual Abilities Evaluation via Med-MIM Benchmark.}
Our constructed Med-MIM Benchmark comprises two parts. 
(a) Held-in part: To comprehensively evaluate four multi-image visual abilities, we construct the held-in Med-MIM Benchmark derived from the Med-MIM instruction dataset,
which includes 2,968 closed-type examples (903, 454, 208, and 1,403 for temporal, reasoning, comparison, co-reference abilities, respectively) and 256 open-type examples (30, 30, 136, and 60, respectively). 
(b) Held-out part: Due to the limited availability of benchmarks for evaluating multi-image visual abilities in the medical domain, we constructed the held-out Med-MIM evaluation benchmark by curating multi-image cases from two complementary sources:  VQA-RAD~\cite{lau2018dataset}, which provides structured QA pairs but grounded in single medical images; ODIR~\cite{li2021benchmark}, containing paired fundus images but lacking annotated QA pairs.
To adapt the standard VQA-RAD benchmark for multi-image scenarios, we first group images based on their modalities, anatomical locations, and semantic coherence with the original QA pairs. We then construct the MIM-RAD by synthesizing composed multi-image instructions from the VQA-RAD dataset, similar to the composed Med-MIM instruction dataset. Additionally, we generate the MIM-ODIR by applying the same procedure used for inherent Med-MIM dataset generation to the ODIR dataset. 
Both MIM-ODIR and MIM-RAD benchmarks are comprised of 300 closed-type cases and 300 open-type cases.
\\
\textbf{Instruction Tuning with Med-MIM Instruction Dataset.}
We conduct instruction tuning from two popular VLMs, LLaVA-Med~\cite{li2024llava} and MANTIS~\cite{Jiang2024MANTISIM}. LLaVA-Med is a successful open-source VLM in the medical domain, while MANTIS has been trained on over 700K multi-natural-image instruction data. 
We fine-tune both VLMs for 3 epochs, running for about 72 hours on 4 NVIDIA A40 GPUs and yielded two specialized models: MIM-LLaVA-Med and Med-Mantis models, both of which excel at understanding medical multi-images.
To accommodate the multi-image and question input, we adopt an interleaved image-text formation and restructure image embeddings as: ``(image \{id\}: <Image> image embeddings </Image>)''.
We employ the negative log-likelihood (NLL) loss function to measure the discrepancy between the predicted next token and the actual next token in the sequence. The specific formula is defined in Eq.~\eqref{eq:loss}, where $K$ represents the total token length of the standard response text. $\mathcal{I}$ consists of interleaved text token sequences ($T_i$) and image token sequences ($I_i$). $p(r^k | \mathcal{I}, q^{1:S}, r^{1:k-1})$ denotes the probability of generating the $k$th token given the interleaved token sequence $\mathcal{I}$, the question token sequence $q^{1:S}$, and the previous tokens in the response sequence $r^{1:k-1}$.
\begin{equation}
\begin{aligned}
    \mathcal{L}_{NLL} = - \sum_{k=1}^K \log p(r^k | \mathcal{I}, q^{1:S}, r^{1:k-1}), \; \mathcal{I} = \textrm{Concat}(T_1, I_1, \dots, T_N, I_N).
\end{aligned}
\label{eq:loss}
\end{equation}

\section{Experimental Results} \label{experiments}
\textbf{Baselines and Metrics.} We select six different SOTA VLMs for comparison including Med-Flamingo-9B~\cite{moor2023med}, Deepseek-VL-7B~\cite{lu2024deepseek}, LLaVA-Med-7B~\cite{li2024llava}, InternVL2-8B~\cite{chen2024internvl}, Mantis-8B~\cite{Jiang2024MANTISIM}, and GPT-4o~\cite{hurst2024gpt}. 
For close-ended (multiple choice questions) evaluation, accuracy is used as the evaluation metric. For open-ended (free-answer questioning) evaluation, the final score is calculated as the average of BERT-recall, BLEU score, and ROUGE-L recall.
\begin{table}[t]
\centering
\caption{Comparison of model performance on the held-in Med-MIM benchmark (temporal understanding, reasoning, comparison, co-reference) and the held-out Med-MIM benchmark (MIM-RAD, MIM-ODIR). Results are expressed as percentages ($\%$). \textbf{C} and \textbf{O} refer to close and open scores, respectively.}
\setlength{\tabcolsep}{0.5pt}
\fontsize{7.9pt}{10pt}\selectfont{
\begin{tabular}{c|cc|cc|cc|cc|cc|cc}
\hline
\multirow{3}*{Models} & \multicolumn{8}{c|}{\textbf{Held-in}} & \multicolumn{4}{c}{\textbf{Held-out}} \\
\cline{2-13}
& \multicolumn{2}{c|}{Temporal} & \multicolumn{2}{c|}{Reasoning} & \multicolumn{2}{c|}{Comparison} & \multicolumn{2}{c|}{Co-reference} & \multicolumn{2}{c|}{MIM-RAD} & \multicolumn{2}{c}{MIM-ODIR} \\ 
\cline{2-13}
& \textbf{C} & \textbf{O} & \textbf{C} & \textbf{O} & \textbf{C} & \textbf{O} & \textbf{C} & \textbf{O} & \textbf{C} & \textbf{O} & \textbf{C} & \textbf{O}
\\
\hline
GPT-4o & 47.95 &  37.28 & 16.96 & 30.18 & 29.81 & 37.20 & 42.30 & 33.81 &\textmd{\bf{69.00}} &\textmd{\bf{39.52}} &\textmd{\bf{45.67}} &34.60\\
\hline
Med-Flamingo-9B &39.65  &24.11  &15.64  &22.01  &32.21  &27.03  &28.72  &27.20  & 4.00  & 28.69  & 16.00 & 30.83\\
deepseek-VL-7B & 29.90 & 36.07 & 14.54 & 34.47 & 32.69 & 34.34 & 29.01 & 31.08 & 17.00 & 31.58 & 21.67  & 44.49 \\
InternVL2-8B & 46.07 & 35.85 &  16.08 & 32.12 & 34.13 & 36.09 & 41.05 & 33.92 &56.33 &\underline{35.62}  &31.67 &41.01 \\
LLaVA-Med-7B & 38.87 & 37.41 & 14.32 & 34.90 & 33.17 & 36.65 & 41.13 & 33.83 &48.67 &34.86  &29.67 &\underline{53.89} \\
Mantis-8B & 68.89 & 37.32 &  15.20 & 24.05 & 38.94 & 31.47 & 35.71 & 22.72 &59.67 &28.08 &23.00 &21.20 \\
\hline
\rowcolor{mygray}
MIM-LLaVA-Med & \underline{74.31} & \underline{44.50} & \underline{22.91} & \underline{41.74} & \underline{50.00} & \bf{40.36} & \underline{46.19} & \underline{43.09} &53.33 &35.08  &\underline{32.33} &\textmd{\bf{55.99}}\\
\rowcolor{mygray}
LLaVA-Med$\uparrow$  & 35.44 & 7.09 & 8.59 & 6.84 & 16.83 & 3.71 & 5.06 & 9.26 &4.66 &0.22 &2.66 &2.10 \\
\hline
\rowcolor{mygray}
Med-Mantis  & \bf{74.86} & \bf{46.87} &  \bf{29.74} & \bf{42.05} & \bf{52.88} & \underline{40.33} & \bf{80.40} & \bf{49.20} &\underline{65.33} &35.35 &30.00 &41.32\\
\rowcolor{mygray}
Mantis$\uparrow$ & 5.97 & 9.55 & 14.54 & 18.00 & 13.94 & 8.86 & 44.69 & 26.48 &5.66 &7.27 &7.00 &20.12\\
\hline
\end{tabular}
}
\label{table:held_in_bench}
\end{table}
\\
\textbf{Main Results on Med-MIM Benchmark.}
\begin{figure}[t]
    \centering
    \begin{tikzpicture}
        \node[anchor=north] (img1) at (-4.0, 0) {\includegraphics[height=0.35\textwidth]{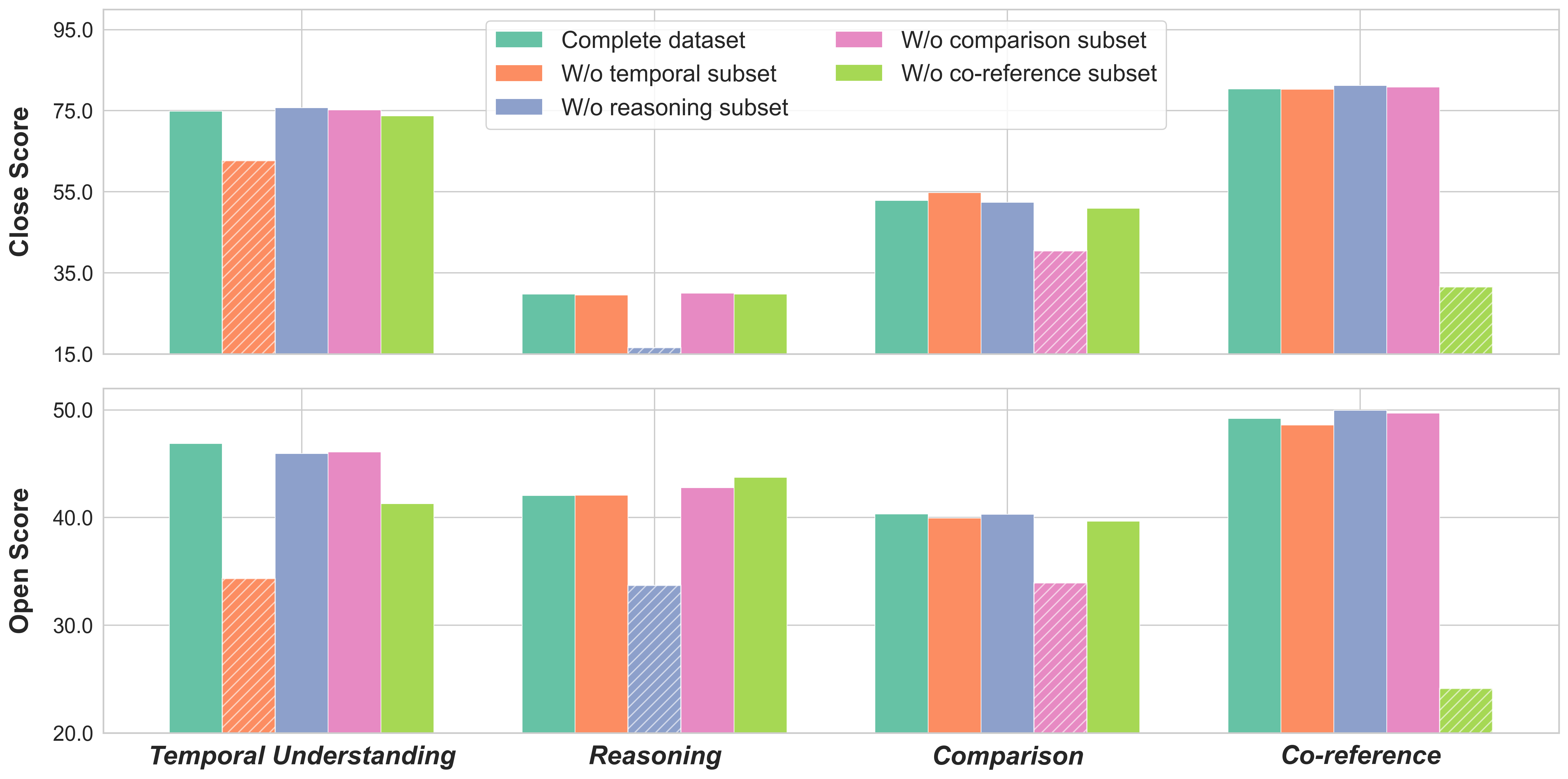}};
        \node[below=-2mm of img1.south] {(a)}; 
        \node[anchor=north] (img2) at (1.6, 0) {\includegraphics[height=0.35\textwidth]{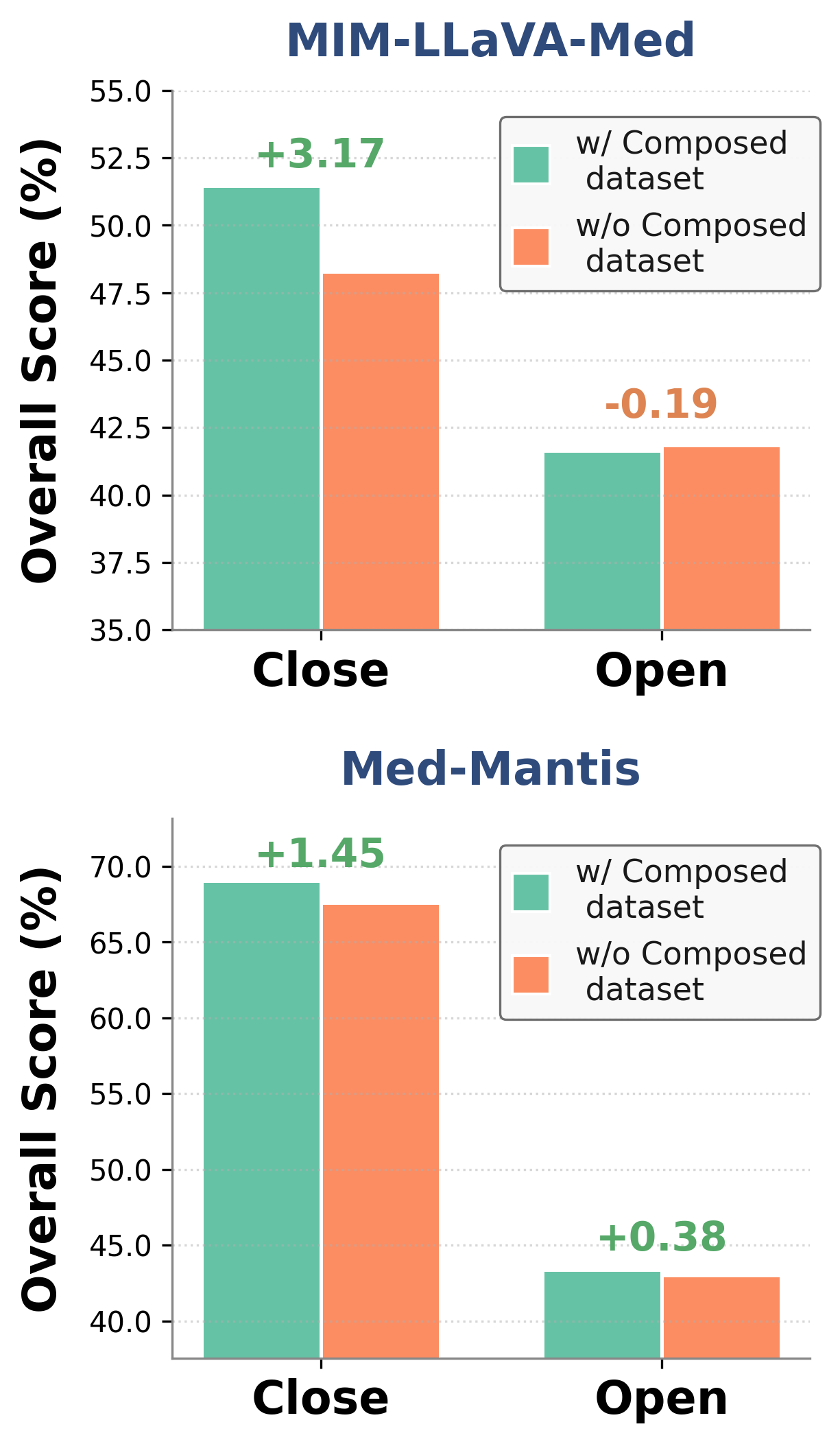}};
        \node[below=-2mm of img2.south] {(b)}; 
    \end{tikzpicture}
    \\
    \begin{tikzpicture}
        \node[anchor=north] (img3) at (-1.5, 0) {\includegraphics[width=0.90\textwidth]{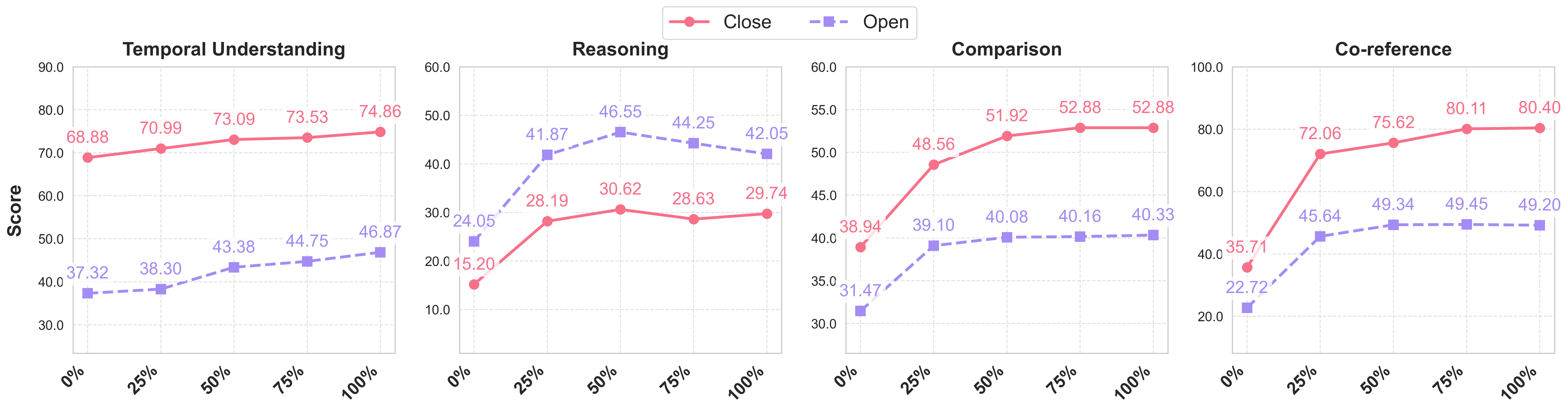}};
        \node[below=-2mm of img3.south] {(c)}; 
    \end{tikzpicture}
    \caption{Ablation results on (a) four visual ability subsets on the Med-Mantis, (b) the performance of Med-Mantis and MIM-LLaVA-Med with and without the composed Med-MIM instruction dataset, and (c) the impact of different sizes of the Med-MIM instruction dataset on the Med-Mantis.}
    \label{fig:ablation_2}
\end{figure}
We evaluate the performance of various LVLMs in both held-in and held-out benchmarks. Table~\ref{table:held_in_bench} presents the corresponding experimental results. The best result for each task is highlighted in \textbf{bold}, while the second-best result is indicated with an \underline{underline}. We present the performance improvement achieved after fine-tuning on the Med-MIM instruction dataset.
For the held-in benchmark, after undergoing supervised fine-tuning on the Med-MIM instruction dataset, both MIM-LLaVA-Med and Med-Mantis show significant improvements across all visual ability subsets. 
Furthermore, compared to other LVLMs, our models secure the top-1 and top-2 performance across all tasks.
We also evaluate the zero-shot performance of models on two held-out benchmarks. 
Experimental results demonstrate that GPT-4o, as one of the most powerful closed-source LVLM, consistently delivers superior performance across both held-out benchmarks.
However, among all open-source LVLMs, our Med-Mantis achieves the highest close-set score on the MIM-RAD benchmark, while MIM-LLaVA-Med attains the best close and open scores on the MIM-ODIR benchmark. Both fine-tuned models exhibit significant performance improvements across all metrics compared to their original versions. This highlights the effectiveness of incorporating Med-MIM instruction dataset, which enables LVLMs to develop advanced medical multi-image inference capabilities.
\\
\textbf{Ablation on Four Visual Abilities.}
To assess the impact of different multi-image visual ability subsets, we perform an ablation study using the Med-Mantis model. We systematically remove subsets corresponding to specific visual abilities from the Med-MIM instruction dataset and fine-tune the model on the remaining data. Fig.\ref{fig:ablation_2}(a) presents the test results for five versions of the Med-Mantis model evaluated on the held-in Med-MIM benchmark. We find that excluding a specific visual ability subset from training leads to a noticeable drop in performance on the corresponding testing set, underscoring the importance of each visual ability subset in the Med-MIM instruction dataset.
\\
\textbf{Influence of the Size to the Med-MIM Instruction Dataset.}
Next, we examine the effect of varying Med-MIM instruction dataset sizes on the performance of the Med-Mantis model. Fig.\ref{fig:ablation_2} (c) illustrates how the model's performance changes when utilizing $0\%$, $25\%$, $50\%$, $75\%$, and $100\%$ of the Med-MIM instruction dataset. The results reveal a consistent improvement in performance as the size of the instruction dataset increases, with notable gains in tasks requiring temporal understanding, comparison, and co-reference visual abilities.
\\
\textbf{Ablation on the Composed Med-MIM Instruction Dataset.}
We investigate the effectiveness of the composed instruction dataset in Fig.\ref{fig:ablation_2} (b). We compare the performance of MIM-LLaVA-Med and Med-Mantis models with and without the composed instruction dataset on the overall held-in benchmark.
The results indicate that the composed instruction dataset significantly improves closed-type QA performance for both models. Furthermore, as Med-Mantis is primarily trained on natural images, fine-tuning with additional composed medical multi-image dataset further boosts its open-type QA performance.
\begin{figure}[!h]
    \centering
	\includegraphics[width=0.90\textwidth]{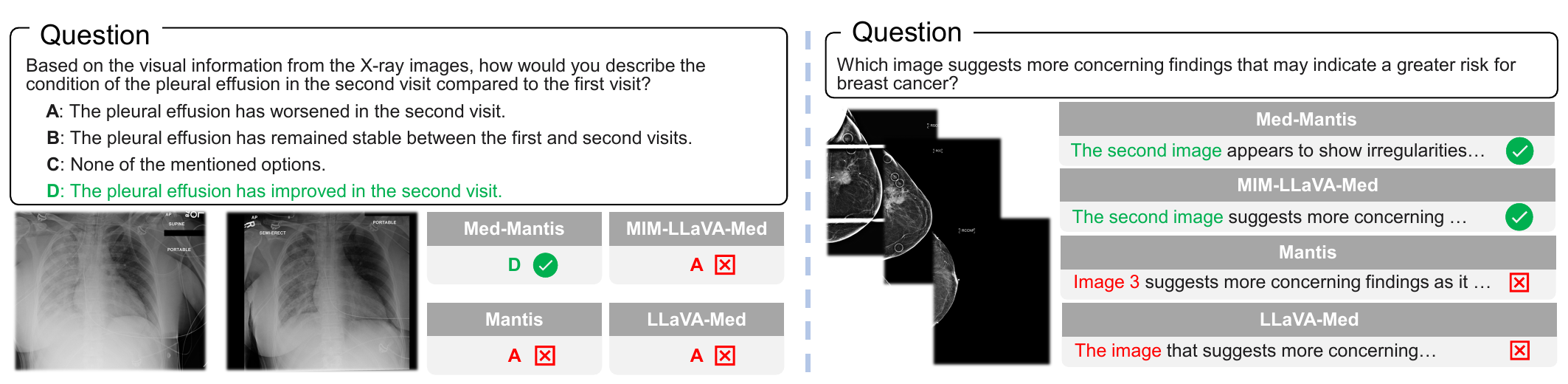}
	\caption{Case studies for LVLMs in medical multi-image understanding.} 
    \label{fig:vis_combine}
\end{figure}
\\
\textbf{Case Study.}
We present two examples in Fig.\ref{fig:vis_combine} from the MS-CXR-T subset (left) and the EMBED subset (right) of the held-in Med-MIM benchmark. The results indicate that Med-Mantis and MIM-LLaVA-Med can produce more accurate response compared with original Mantis and LLaVA-Med models.

\section{Conclusion}
In this work, we present the Med-MIM instruction dataset, designed to enhance medical multi-image visual abilities of LVLMs. Leveraging this dataset, we fine-tune two models, Med-Mantis and MIM-LLaVA-Med, both of which demonstrate superior performance in answering medical questions involving multiple images. To further assess the multi-image understanding of LVLMs, we construct the Med-MIM benchmark. Experimental results comparing our models to six popular LVLMs reveal that fine-tuning with the Med-MIM instruction dataset significantly enhances multi-image visual abilities in the medical domain.

%
%

%
%
%
\bibliographystyle{splncs04}
\bibliography{refs}
%






\end{document}